\documentclass[12pt,journal,compsoc,onecolumn]{IEEEtran}

\usepackage{amsmath,graphicx,multirow}
\usepackage{amsthm} 
\usepackage{amssymb}
\usepackage{graphicx}
\usepackage{algorithm2e}
\usepackage{dsfont}
\usepackage{algorithmic}
\usepackage{color} 
\usepackage{booktabs}

\title{Image Classification  with Deep Reinforcement Active Learning\thanks{This work is supported by grants from the National Natural Science Foundation of China (No.~62272422, U22B2051).}}
\author{Mingyuan Jiu$^{1,2,3}$ \ \ \and \ \ Xuguang Song$^{1}$ \ \ \and \ \ Hichem Sahbi$^{4}$ \ \ \and \ \ Shupan Li$^{1,2,3}$ \ \ \and \ \ \\ Yan Chen$^{5}$ \ \ \and \ \  Wei Guo$^{5}$ \ \ \and \ \ Lihua Guo$^{5}$ \ \ \and \ \ Mingliang Xu$^{1,2,3}$ \\
\textit{\small  $^1$School of Computer and Artificial Intelligence, Zhengzhou University, China} \and \\
	$^2$\textit{\small Engineering Research Center of Intelligent Swarm Systems, Ministry of Education, China} \and \\
	$^3$\textit{\small National Supercomputing Center in Zhengzhou, China} \and \\
	$^4$\textit{\small Sorbonne University, CNRS, LIP6, F-75005, Paris, France} \and \\
	$^5$\textit{\small Luoyang float glass group Co.,Ltd; State Key Laboratory of Advanced Technology for
	Float Glass, China}
}

\begin{document}
 \maketitle
\begin{abstract}
  Deep learning is  currently reaching outstanding performances on different tasks, including image classification, especially when using large neural networks. The success of these models is tributary to the availability of large collections of labeled training data. In many real-world scenarios, labeled data are scarce, and their hand-labeling is time, effort and cost demanding. Active learning is an alternative paradigm that mitigates the effort in hand-labeling data, where only a small fraction is iteratively selected from a large pool of unlabeled data, and annotated by an expert (a.k.a oracle), and eventually used to update the learning models. However, existing active learning solutions are dependent on handcrafted strategies that may fail in highly variable learning environments (datasets, scenarios, etc). In this work, we devise an adaptive active learning method based on Markov Decision Process (MDP). Our framework leverages deep reinforcement learning and active learning  together with a Deep Deterministic Policy Gradient (DDPG) in order to dynamically   adapt sample selection strategies to  the oracle's feedback and the learning environment.   Extensive experiments conducted on three different image classification benchmarks show superior performances against several existing active learning strategies. 

\noindent {\bf keywords.} {Active learning, Deep reinforcement learning, Image classification.}
\end{abstract}

\section{Introduction}
With the emergence  of deep learning,  image classification is currently reaching unprecedented performances. Deep neural networks, including convolutional neural networks (CNN)~\cite{Chen2021ReviewOI} and transformers~\cite{Vaswaninips17}, have achieved a major leap in image classification by automatically learning discriminative features from images. However, traditional deep learning solutions heavily rely on large collections of training data whose hand-labeling is dilatory.   Hence,  designing  {\it label-frugal} models while guaranteeing their accuracy is currently one of the major trends  in deep learning.   Active learning --- as an effective paradigm --- addresses  the aforementioned challenge by  selecting  {\it only} the most informative  samples~\cite{10.1016/j.eswa.2017.05.046} from a large collection of unlabeled data using different strategies.  The  informative samples  are afterwards used to probe the oracle about their labels and to train classification models while maintaining equivalent performances w.r.t. their non-frugal counterparts.  Hence, the design of appropriate sample selection strategies is critical in order to guarantee equivalent performances. Early strategies rely on  uncertainty~\cite{caramalau2021visual,10.1145/3534932,Tang2023TransformerbasedML}, and representativity~\cite{5611590,Gissin2019DiscriminativeAL},  as well as diversity~\cite{xie2021towards}; nonetheless, these staple strategies  are mostly  handcrafted. Other more sophisticated criteria are based on the optimization of well-designed loss functions~\cite{Dasgupta2004AnalysisOA, sahbiicpr22}. However, as training environments may evolve, the above mentioned strategies may not always select the most informative samples resulting into ineffective use of the labeling budgets and thereby suboptimal classification performances.  \\

\indent In this paper, we devise a novel active learning model based on reinforcement learning. The proposed method, dubbed as Deep Reinforcement Active Learning (DRAL), casts active learning as a Markov Decision Process (MDP). The latter aims at selecting the most informative samples in order to {\it frugally} query the oracle about their labels. Prior to achieve sample selection, an uncertainty criterion is first adopted for sample sorting. Then, reinforcement learning is performed based on the oracle's feedback  resulting into highly effective (dynamic) selection strategies compared to handcrafted ones. Besides, an actor-critic module is leveraged in order to design the sample selection strategy, and the model is accordingly optimized using a Deep Deterministic Policy Gradient (DDPG) procedure~\cite{lillicrap2015continuous}. Considering the aforementioned issues, the main contributions of this work include (i) a new active learning model based on MDP and reinforcement learning, (ii) a novel design of environment-dependent sample selection strategies, and  (iii) extensive experiments including different benchmarks showing the out-performance of our proposed method against the related work.
\section{Related Work}
Many active learning strategies exist in the literature; some of them are heuristics whilst others are more principled and rely on well-designed objective functions. This includes uncertainty-based approaches which select samples with the most uncertain prediction results defined by their probability distribution~\cite{caramalau2021visual}. Representativity-based methods proceed differently and select samples as representatives of the underlying unlabeled data distributions; in other words, relevant samples are selected according to feature distributions or other indicators~\cite{gissin2019discriminative}. These criteria usually prefer samples farthest away from the labeled ones or those with the largest difference from the already labeled samples in order to ensure enough variability in the selected data~\cite{10.1145/3324884.3416621}. Other methods are iterative and proceed by minimizing a constrained  objective function mixing diversity and representativity as well as uncertainty~\cite{deschamps2022reinforcement}.\\

With the resurgence of neural networks, deep reinforcement learning has gained a particular attention due to its ability to learn actions (or strategies) in highly dynamic environments. In active learning, Fang et al.~\cite{fang2017learning} redefine sample selection heuristics using reinforcement learning whereas Haussmann et al.~\cite{haussmann2019deep} propose a reinforced active learning algorithm which relies on Bayesian neural networks for sample selection. Liu et al.~\cite{2019Deep} adopt a similar deep reinforcement active learning idea while Sun et al.~\cite{8901911} leverage deep convolutional neural networks to extract image features as the "state" of reinforcement learning, and adopt the deep Q-learning algorithm to train Q-networks. The outputs of the latter are used to decide whether a given sample  is informative for further labeling and model retraining.   Finally,  Wang et al.~\cite{slade2022deep} consider active learning as a Markov decision process where  a deep deterministic strategy gradient algorithm is used to train the model based on a reinforcement learning algorithm and an Actor-Critic architecture.  However,  the  approach in \cite{slade2022deep} is more suitable for small and mid-scale datasets,  and computationally expensive on larger datasets.   In contrast,  our method,  described  in the subsequent sections,   overcomes this computational issue  by sorting and preselecting only a small fraction of unlabeled data using a relevant uncertainty criterion,  prior to feed the preselected data to an actor network.

\def\S{{\cal S}}
\def\A{{\cal A}}
\def\x{{\bf x}}
\def\X{{\bf X}}

\section{Methodology}
\subsection{Method Overview}
\begin{figure}[t]
	\centering
	\includegraphics[scale=1.15]{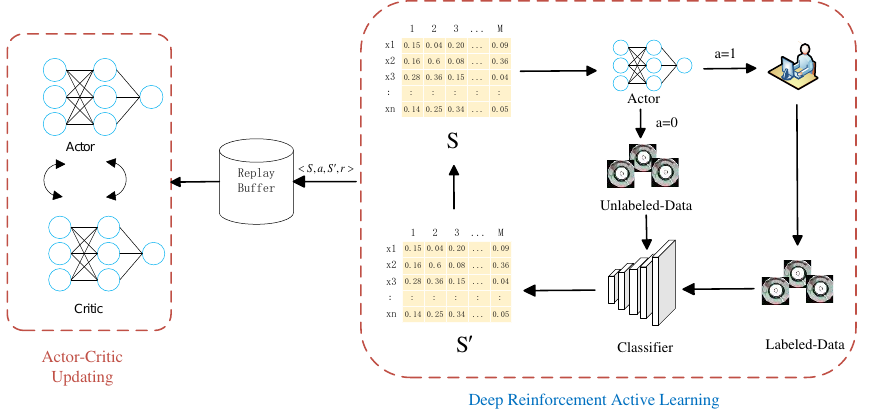}
	\caption{The proposed Deep Reinforcement Active Learning framework.} 
	\label{Fig.main1} 
\end{figure}

Fig.~\ref{Fig.main1} illustrates our DRAL model which casts active learning as a Markov decision process. Our model is composed of {\it a state set $\cal S$, action set $\cal A$, a reward function $r: \S \times \A \mapsto \mathbb{R}$, a transition function $q: \S \times \A \mapsto \S$ and a global predefined labeling budget $B$}. Our method first reorders all unlabeled samples using an uncertainty measure. Then, it selects the top $n$ samples (according to this measure) and uses the features extracted by a deep neural network (see details subsequently) as a global state of the model. The state is afterwards fed to an actor network, where its output action is either 1 or 0,  suggesting that the oracle should label these samples (or not); {\it only if} the actor's outputs equate 1,   the samples are labeled by the oracle, and used in the subsequent reward estimation. This reward is defined as the difference between the accuracy of the classifier after and before selecting the samples; a positive difference stands for a positive impact of the selected samples, and vice-versa. Only when the difference is positive, the selected samples are added to the pool of labeled training data in order to further retrain the classifier. This procedure, shown in Fig.~\ref{Fig.main1}, is repeated till exhausting the labeling budget. 

\subsection{Deep Reinforcement Active Learning}
Our proposed DRAL relies on actor-critic networks as the core components of our model. The actor network is used to select the most informative samples from the unlabeled set based on a current state $S_t \in \S$ and the learned strategies (i.e., actions); here the subscript $t$ stands for the current active learning iteration. The critic network decides whether the learned actions improve performances or not. Both networks are trained as a part of our deep reinforcement learning framework which allows reaching better sample selection strategies as shown subsequently and also later in experiments. 
\subsubsection{States}
The global state of our model is obtained by feeding all the unlabeled samples to a neural network classifier that extracts their features. The latter are afterwards ranked according to an uncertainty criterion based on the classifier margin.  Formally, the current state $S_t \in \S$ is obtained as  
\begin{equation}\label{eq000}
S_t = \X_u^n(t),
\end{equation}
being $\X_u^n(t)$ the matrix formed by concatenating the top $n$ ranked unlabeled samples according to the  margin uncertainty of the classifier  at the $t$-th iteration of active learning.
\subsubsection{Actions}
We feed the top $n$ samples in $S_t$ to an actor network in order to obtain an action for each entry in $S_t$. We adopt, as actor, a network with two convolutional layers and one pooling layer, followed by three fully connected layers, using \textit{tanh} activations to map the results to $[-1, 1]$. For negative activations,  the corresponding actions are set to $0$ corresponding to discarded samples while positive activations stand for selected samples. The latter are used to probe the oracle about their labels, and labeled data are added to a global training set.   Considering $\theta_a$ as the parameters of the actor network, designing relevant sample selection strategies translates into learning a policy $\pi(S_t; \theta_a)$ that generates an action $a \in {\cal A}$ for each element in $S_t$.

\subsubsection{Transitions}
This function $q$ defines the subsequent state $S_{t+1}$ as a set of $n$ unlabeled samples $\X_u^n(t+1)$ with highest margin uncertainty (as described earlier and also in Eq.~\ref{eq000}). It's worth noticing that we do not obtain $b$ samples at once and submit them to the oracle for labeling (where $b$ can be understood as a fixed budget that triggers manual labeling from the oracle). Hence, prior to reach $b$, we make use of the ranked sample sequence delivered by the classifier, as well as the extracted features. Once $b$ is reached, the $b$ samples are labeled and fed to the classifier for further retraining.

\subsubsection{Rewards}
The design of the reward functions is crucial for the success of reinforcement learning. This function is used to assess the relevance of action-state pairs, and also to guide the actor network to choose relevant samples for further labeling and retraining. However, in many reinforcement learning algorithms, rewards are provided only after the action ends, leading to delayed feedback during the learning process. This may  affect  the relevance  of  the rewards  w.r.t.   past actions.  In order to circumvent this issue, we consider  a more relevant  reward that captures  changes in the  classifier's  accuracy at each reinforcement learning iteration $t$ as
\begin{equation}
r(S_t,a) = Acc(\phi_t)-Acc(\phi_{t-1}),
\end{equation}
here  $Acc$ denotes the accuracy of the  classifier  $\phi_t$ when $a=1$.  In what follows, we rewrite $S_t$, $S_{t+1}$ simply as $S$, $S'$ respectively. Our objective is to maximize the reward through interaction with the environment using a $Q$-value function. The Bellman equation is used to establish the relationship between states and actions as
\begin{equation}
Q(S,a;\theta_c) = \mathbb{E}_\pi\big[\gamma{Q}(S',\pi({S';\theta_a});\theta_c)+r(S,a)\big],
\label{eq:bellman}
\end{equation}
\noindent being $\theta_c$ the parameters of the critic network that approximates the $Q$-value function, and $\gamma$ a decay factor that balances between future and immediate rewards. Inspired by deep $Q$-learning~\cite{Mnih2015HumanlevelCT}, we design a greedy policy for the actor network by solving a maximization problem
\begin{equation}
\mathop{\max}_{\theta_a} \ Q(S,\pi(S;\theta_a);\theta_c).
\end{equation}
Then we define $\tilde{\text{Q}}(S,a;\theta_c) = \gamma{Q(S',\pi(S';\theta_a);\theta_c)}+r(S,a)$, and learn the critic network by solving the following problem
\begin{equation}
\mathop{\min}_{\theta_c}\big(\tilde{\text{Q}}(S,a;\theta_c)-Q(S,a;\theta_c)\big)^2.
\label{eq:criticproblem}
\end{equation}

\subsubsection{Budget}
We fix $B$ as a global labeling budget~\cite{fang2017learning}; put differently, it corresponds to the total number of samples submitted to the oracle for labeling. This budget $B$ is also proportional to the max number of mini-batches in each training epoch. $B$ is reached by appending multiple $b$ sample sets in order to form the global training set as described earlier. 

\subsubsection{Training with target networks} In order to achieve better performance, we follow~\cite{10.1007/978-3-030-73197-7_36} which employs a separate target network in order to evaluate $\tilde{\text{Q}}(S,a;\theta_c)$. According to Eq.~\eqref{eq:criticproblem}, $\pi(S';\theta_a)$ depends on the actor's output, $\tilde{\text{Q}}(S,a;\theta_c)$ depends on the
subsequent state $S'$, and $(S',\pi(S',\theta_a))$ is evaluated by the critic. Therefore, we adopt a separate target actor network parameterized by $\theta_{a'}$ and a separate target critic network parameterized by $\theta_{c'}$ in order to evaluate  $\tilde{\text{Q}}(S,a;\theta_c)$.
Therefore, Eq.~\eqref{eq:criticproblem} is rewritten as
\begin{equation}
\mathop{\min}_{\theta_c}\big(\gamma Q'(S',\pi'(S';\theta_{a'});\theta_{c'})+r(S,a)-Q(S,a;\theta_c)\big)^2,
\label{eq:sepactorcritic}
\end{equation}
where $\pi'(.^.,\theta_{a'})$ is the policy from the target actor, and $Q'(.,.^.,\theta_{c'})$ is the function from the target critic. Deep deterministic policy gradient algorithm (DDPG)~\cite{10.1007/978-3-030-73197-7_36} is applied to solve the optimization problem in Eq.~\eqref{eq:sepactorcritic}. At the end of each epoch, the target actor and critic are updated by
\begin{equation}
\theta_{a'} := \lambda \theta_a + (1-\lambda)\theta_{a'}, \qquad  \theta_{c'} := \lambda \theta_c + (1-\lambda)\theta_{c'},
\end{equation}
where $\lambda \in (0,1)$ is a trade-off parameter.  During training, we store each obtained transition $(S,a,S',r)$ in a replay buffer. This allows  repeatedly  utilizing  these experiences for learning, suppressing data correlations, and thus improving learning efficiency and stability. Afterwards,  we select mini-batches of transitions to update both the actor and critic networks.

\begin{figure*}[t]
	\centering
	\includegraphics[width=0.99\textwidth]{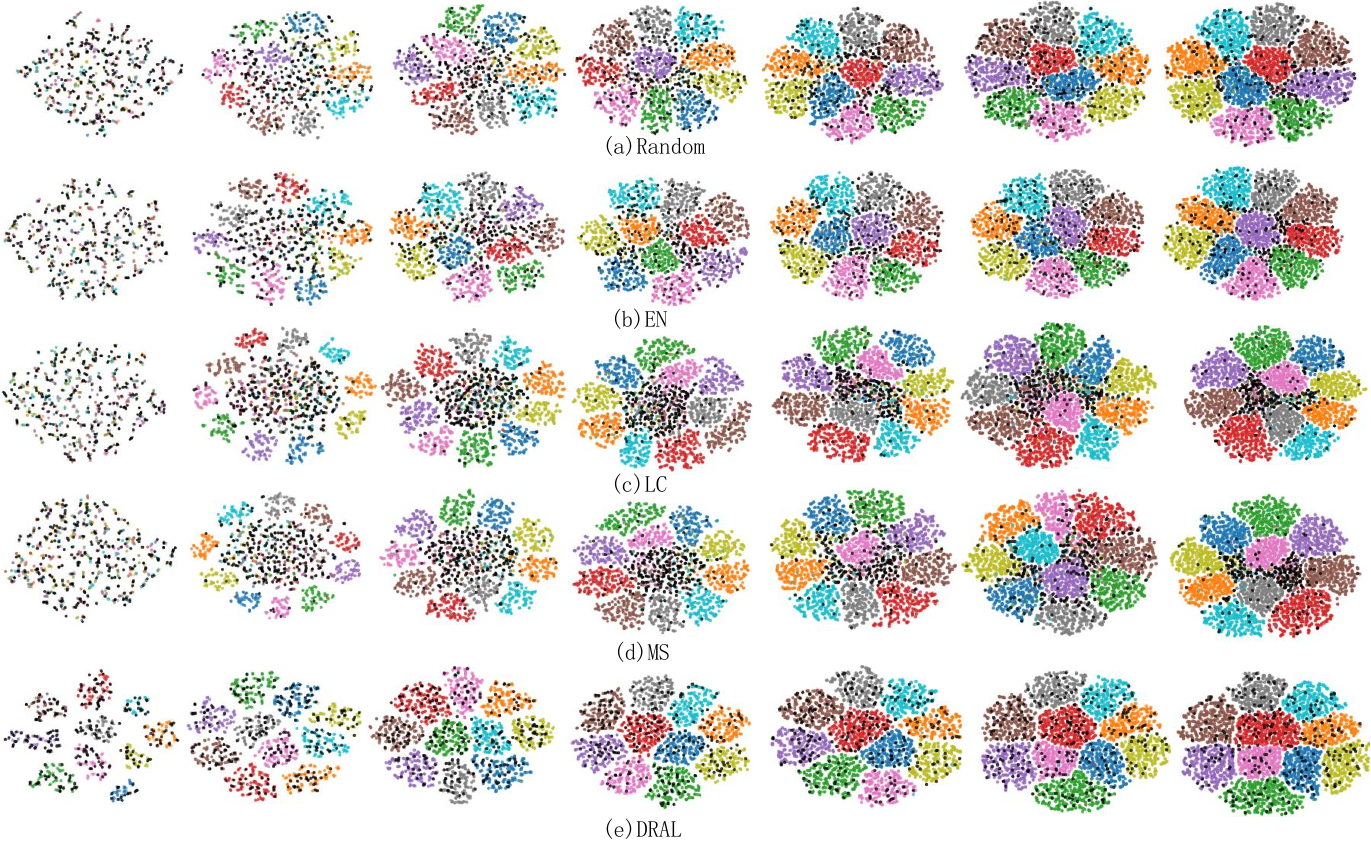}
	\vspace{-0.3cm}
	\caption{Comparison of visualization of the two-dimensional distribution of the selected samples by different methods using t-SNE on the CIFAR-10 dataset. Each row shows the results obtained by one method, each column stands for selected samples in each iteration. In each iteration, the points with the same color have the same label and black points correspond to the selected samples.}
	\label{fig:distribution}
\end{figure*}

\section{Experiments}

\subsection{Experimental Settings}
We evaluate the proposed method on CIFAR-10, SVHN, and Fashion-Mnist benchmarks. They are widely used to evaluate the performance of active learning for image classification. CIFAR-10  includes 60k  images, divided into a training set with 50k  images and a test set with 10k images.  SVHN  consists of a training set with 73,257 images and a test set with 26,032 images extracted from Google Street View. Similarly,  Fashion-MNIST is also divided into a training set with 60k images and a test set with 10k  images.  All these sets include ten categories.\\

Due to its relatively high accuracy and better training stability, ResNet-18~\cite{7780459} is chosen as the learner to evaluate the model on the final selected samples. For the classifier used in sorting samples during the selection, a traditional CNN is adopted. The initial learning rate is set to  0.01 and the batch size is 128. Stochastic Gradient Descent (SGD) with a weight decay $5 \times {10}^{-4}$ and a momentum of 0.9~\cite{caramalau2021sequential} is applied. The maximal number of training epochs is 100. All the experiments are implemented on the PyTorch platform~\cite{paszke2019pytorch}.\\

The actor and critic networks with the same structure are adopted, consisting of five fully connected layers. The Adam optimizer is used and the learning rate and the delay factor $\gamma$ are respectively set to  0.0001 and 0.99. The trade-off parameter is set to $\lambda = 0.01$. The size of the replay buffer is set to  3000. Random sampling is used only when the storage length exceeds 128, and the size of randomly selected samples  is 64 at each iteration $t$. The small budget $b$ is set to 100.

\begin{table}[htbp]
	\caption{Results on the CIFAR-10 dataset w.r.t. different sample sizes}
	\label{tab:rescifar}
	\centering
	\vspace{-0.2cm}
	\begin{tabular}{cccccccc}
		\toprule
		Sizes & 1000 & 2000 & 3000 & 4000 & 5000 & 6000 & 7000 \\
		\midrule
		RANDOM & 50.04 & 57.00 & 60.84 & 64.44 & 66.27 & 67.36 & 69.13\\
		EN~\cite{Shannon1948AMT} & 49.36 & 57.83 & 61.88 & 64.75 & 67.03 & 67.94 & 69.16\\
		LC~\cite{Settles2009ActiveLL} & 50.60 & 57.80 & 61.60 & 65.01 & 66.58 & 68.21 & 69.73\\
		MS~\cite{10.5555/647967.741626} & 50.65 & 58.15 & 62.53 & 65.77 & 67.46 & 68.84 & 70.39\\
		\hline
		\textbf{DRAL} & \textbf{51.89} & \textbf{59.01} & \textbf{62.99} & \textbf{66.25} & \textbf{68.20} & \textbf{69.45} & \textbf{71.01}\\
		\bottomrule
	\end{tabular}
\end{table}

\begin{table}[tbp]
	\caption{Results on the SVHN dataset w.r.t. different sample sizes}
	\label{tab:ressvhn}
	\centering
	\vspace{-0.2cm}
	\begin{tabular}{cccccccc}
		\toprule
		Sizes & 1000 & 2000 & 3000 & 4000 & 5000 & 6000 & 7000 \\
		\midrule
		RANDOM & 73.73 & 85.23 & 88.16 & 90.05 & 90.06 & 91.35 & 91.61\\
		EN~\cite{Shannon1948AMT} & 72.80 & 85.38 & 88.32 & 90.38 & 91.13 & 92.57 & 93.43\\
		LC~\cite{Settles2009ActiveLL} & 74.51 & 85.86 & 88.86 & 91.20 & 91.24 & 92.68 & 93.56\\
		MS~\cite{10.5555/647967.741626} & 74.93 & 86.17 & 89.12 & 91.25 & 91.42 & 93.00 & 93.40\\
		\hline
		\textbf{DRAL} & \textbf{75.08} & \textbf{86.85} & \textbf{90.31} & \textbf{91.81} & \textbf{92.04} & \textbf{93.34} & \textbf{93.90}\\
		\hline
		
	\end{tabular}
\end{table}

\begin{table}[tbp]
	\caption{Results on Fashion-Mnist w.r.t. different sample sizes}
	\label{tab:resmnist}
	\centering
	\vspace{-0.2cm}
	\begin{tabular}{cccccccc}
		\toprule
		Sizes & 1000 & 2000 & 3000 & 4000 & 5000 & 6000 & 7000 \\
		\midrule
		RANDOM & 79.13 & 82.70 & 84.46 & 85.16 & 86.06 & 86.72 & 86.93\\
		EN~\cite{Shannon1948AMT} & 79.28 & 82.83 & 84.37 & 86.18 & 87.28 & 87.46 & 88.76\\
		LC~\cite{Settles2009ActiveLL} & 77.93 & 83.17 & 85.91 & 86.62 & 87.37 & 88.15 & 88.80\\
		MS~\cite{10.5555/647967.741626} & 78.74 & 84.02 & 85.68 & 87.10 & 88.08 & 88.11 & 88.85\\
		\hline
		\textbf{DRAL} & \textbf{80.45} & \textbf{85.12} & \textbf{86.71} & \textbf{87.57} & \textbf{88.44} & \textbf{88.72} & \textbf{89.11}\\
		\hline
	\end{tabular}
\end{table}

\subsection{Results and Discussion}
We compare the proposed method with several existing active learning methods, random selection (RANDOM), least confidence (LC)~\cite{Settles2009ActiveLL}, margin sampling (MS)~\cite{10.5555/647967.741626}, and entropy (EN)~\cite{Shannon1948AMT}. For each method, we first randomly select 1000 images from the unlabeled samples $\mathcal{D}_u$ as a start set. These images serve as the labeled dataset $\mathcal{D}_l$ and are used to train an initial classifier. In the training process, when the number of selected samples for each selection reaches $b$, they are added to $\mathcal{D}_l$ to form a new $\mathcal{D}_{l+{new}}$ and the classifier is retrained in the subsequent selection. This process is repeated iteratively till reaching the global budget ${B}$.\\

Tab.~\ref{tab:rescifar}, \ref{tab:ressvhn}, \ref{tab:resmnist} respectively show the comparison results of the proposed method with several other active learning methods on the CIFAR-10, Fashion-MNIST, and SVHN datasets. It is clear that when selecting the same number of samples, our proposed method significantly outperforms the other active learning methods, which validates the effectiveness of the dynamic sample selection strategy.
We also visualize the distributions of the selected samples generated by different methods on the CIFAR-10 dataset by using t-SNE method~\cite{2008Visualizingtsne} in Fig.~\ref{fig:distribution}. In each iteration, 1000 new images (displayed as black dots) are selected for manual annotation, while the other colored dots indicate the distribution of different classes after retraining the model with the selected samples. From the visualizations, we clearly observe that our method selects samples that are more evenly distributed, and  it distinguishes different classes with clear boundaries. This indicates that the proposed method is able to capture a broader range of sample information, resulting in a better representative selection.

\section{Conclusion}
This paper introduces a novel Deep Reinforcement Active Learning (DRAL) approach that casts sample selection as a Markov decision process. The strength of DRAL resides in its ability to overcome the limitations of handcrafted sample selection strategies (widely used in conventional active learning) using actor-critic networks that allow designing sample selection strategies well suited to dynamic environments. Deep deterministic policy gradient algorithm is also leveraged in order to train these models. Extensive experiments, involving three image classification benchmarks, show a consistent gain of our proposed DRAL method against the related work.

\end{document}